\documentclass[runningheads,a4paper]{llncs}
\usepackage{paralist}

\usepackage{graphicx}

\usepackage{siunitx}

\usepackage[table]{xcolor}
\usepackage{booktabs}
\usepackage{multirow}

\usepackage[hidelinks]{hyperref}
\newcommand\rurl[1]{%
\texttt{\href{http://#1}{\nolinkurl{#1}}}%
}

\usepackage{cleveref}
\crefname{table}{Tab.}{Tabs.}
\crefname{figure}{Fig.}{Figs.}
\crefname{section}{Sec.}{Secs.}
\crefname{equation}{Eq.}{Eqs.}

\usepackage[labelformat=simple]{subcaption}

\usepackage{fontawesome}
\usepackage{pifont}

\usepackage{copyright}

\begin{document}

\mainmatter
\title{
\textit{What you see is what you get}: Experience ranking with deep neural dataset-to-dataset similarity for topological localisation
}
\titlerunning{ What you see is what you get }
\author{
Matthew Gadd\thanks{
Supported by EPSRC Programme Grant ``From Sensing to Collaboration'' (EP/V000748/1).
}, Benjamin Ramtoula\thanks{Suppported by EPSRC Centre for Doctoral Training in Autonomous Intelligent Machines and Systems (EP/S024050/1), and Oxa.}, Daniele De Martini, Paul Newman
}
\authorrunning{ M. Gadd \textit{et al.} }
\institute{ Mobile Robotics Group, University of Oxford\\
\faEnvelope~\texttt{mattgadd@robots.ox.ac.uk}\\
\faGithub~\rurl{github.com/mttgdd/vdna-experience-selection}
}
\maketitle

\begin{abstract}
Recalling the most relevant visual memories for localisation
or understanding \textit{a priori} the likely outcome of localisation effort against a particular visual memory is useful for efficient and robust visual navigation.
Solutions to this problem should be divorced from performance appraisal against ground truth -- as this is not available at run-time -- and should ideally be based on generalisable environmental observations.
For this, we propose applying the recently developed \textit{Visual DNA} as a highly scalable tool for comparing datasets of images -- in this work, sequences of past (map) and live experiences.
In the case of localisation, important dataset differences impacting performance are modes of appearance change, including weather, lighting, and season.
Specifically, for any deep architecture which is used for place recognition by matching feature volumes at a particular layer, we use distribution measures to compare neuron-wise activation statistics between live images and multiple previously recorded past experiences, with a potentially large seasonal (winter/summer) or time of day (day/night) shift.
We find that differences in these statistics correlate to performance when localising using a past experience with the same appearance gap.
We validate our approach over the \textit{Nordland} cross-season dataset as well as data from Oxford's \textit{University Parks} with lighting and mild seasonal change, showing excellent ability of our system to rank actual localisation performance across candidate experiences.
\end{abstract}
\begin{keywords}
Localisation, Deep Learning, Autonomous Vehicles
\end{keywords}

\copyrightnotice

\section{Introduction}
\label{sec:introduction}

\textit{Topological localisation and place recognition} are about matching sequences of images which can be viewed as image datasets under domain shift.
A useful \textit{measure of domain gap} will enable \textit{localisation performance prediction and experience ranking} to prioritise memories for localisation.
As will be shown in~\cref{fig:spring-diff-mats} in~\cref{sec:optimal}, single-experience selection either achieves the best performance or is closest to the performance of a multi-experience map.

In more detail, \textit{visual topological localisation} in a teach-and-repeat setup is an effective way to localise a robot after having built maps of an environment~\cite{furgale2010visual,dequaire2016off,krajnik2017image,warren2018there}.
This involves retrieving from a map the most similar images to the robot's current observations.
This place recognition competency, therefore, needs to be robust to variations in appearance due to season, lighting, etc~\cite{lowry2015visual}.
Modern approaches learn stable representations of places from images through neural networks, first achieved across severe appearance change in~\cite{gomez1505training}, later achieved at vast scale in~\cite{chen2017deep}, and with state-of-the-art mixing of features across network layers~\cite{ali2023mixvpr}.
However, it was shown in~\cite{sunderhauf2015performance} that, depending on how networks are trained, late network layers are robust to viewpoint variation but may be sensitive to extreme appearance change.
Looking more closely within neural network layers, \cite{hausler2018feature} employ ``feature map filtering'' to remove feature maps that exhibit variance in their activation when the appearance of a scene changes over time.

Here, appearance variation can be considered a case of \textit{domain shift}.
While learned place-recognition work often targets representations invariant to condition changes, recent work has begun to measure the severity of those differences between sets of images.
For example, \textit{Visual DNA}~\cite{ramtoula2023cvpr} represents image collections by the distributions of neuron activations in pre-trained network architectures (which we term \texttt{vdna}) and aggregates neuron-wise distribution comparisons to measure dataset differences over many levels of features.

We apply this to the domain-shifted place recognition problem by performing \textit{experience ranking} to improve localisation.
Instead of training models fully robust to appearance change, in this experimental study we are interested in selecting \textit{a priori}, among a pool of potential experience candidates, the one with the appearance that maximises such models' performances.
We ask the question: \textit{can we improve efficiency and accuracy by selecting the most relevant maps for a given deployment condition, a-priori}?
For this, we relate dataset-to-dataset similarity to localisation performance.
In particular, we are interested in \textit{low-level illumination and high-level seasonal shift}.
We show that, while simple pixel-intensity measures are a proxy for the former, ours is generally a better prior belief in the relevance of visual experiences.

Moreover, thanks to the \textit{Visual DNA} mechanism, despite being based on large neural networks, we have measured our method performing experience selection with $100$ live images in \SI{16}{\second} or less using a robot's onboard CPU only, meaning experience-selection is feasible more quickly than a robot's immediate surroundings would significantly change if travelling at a reasonable pace.  

\section{Related Work}
\label{sec:related_work}

In the area of \textit{experience selection}, related work includes~\cite{linegar2015work,mactavish2017visual,gadd2016checkout}.
Linegar \textit{et al.}~\cite{linegar2015work} use a probabilistic formulation over the recent localisation history to predict which nodes -- and therefore experiences -- are currently relevant for localisation.
Gadd and Newman~\cite{gadd2016checkout} develop policies for disseminating visual experiences amongst a fleet of centrally communicating vehicles.
MacTavish \textit{et al.}~\cite{mactavish2017visual} compare a live image's bag-of-words to a locally constructed vocabulary of visual features and in~\cite{mactavish2018selective} perform landmark-level recommendations.
None of these rely on robust representations learned by neural networks, as ours does.

Tu \textit{et al.}~\cite{tu2023bag} assess training-set suitability and test-set difficulty, performing dataset vectorisation by projecting image features onto a codebook obtained by clustering.
Ramtoula \textit{et al.}~\cite{ramtoula2023cvpr} predict test-set performance from training dataset similarity in semantic segmentation.
None of these methods investigates localisation or place recognition as a task, as we do.

\section{Technical Approach}
\label{sec:method}

\Cref{fig:vdna-overview,fig:dna-pr-system} show our system.
In brief, we represent all images in an experience or sequence by a \texttt{vdna}, and \texttt{vdnas} over sequences representative of one type of appearance change can be compared to predict the place recognition performance between further sequences from those conditions.

\subsection{Topological localisation system}
\label{sec:topological_localisation_system}

\begin{figure}[t]
\centering
\vspace{-10pt}
\includegraphics[width=0.9\textwidth]{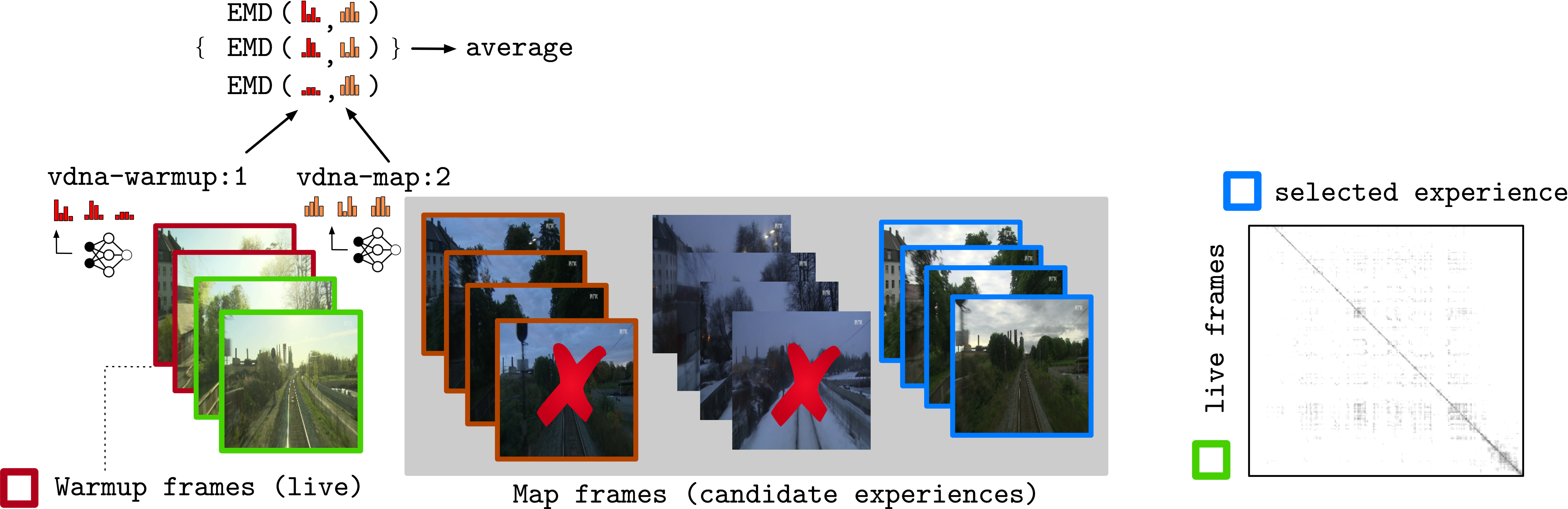}
\caption{
\textit{What you see is what you get} system overview.
The \texttt{map} frames (grey box) are images to localise to, from some prior traversal of the environment, with each \texttt{map} sequence referred to as an ``experience''. 
\texttt{Vdnas} are generated to represent the image domains of recorded experiences to store them alongside the map (e.g. orange bar chart).
Online, one \texttt{candidate experience} is selected by finding the most similar \texttt{vdnas} to a brief window of \texttt{warmup} frames (red frame) before localising the \texttt{live experience} (green frames). 
}
\label{fig:dna-pr-system}
\vspace{-15pt}
\end{figure}

We record multiple experiences of traversals of a chosen trajectory, complete with images and robot poses -- e.g. the sequences in the grey shaded area of~\cref{fig:dna-pr-system}.

During deployment, the robot is localised by performing image retrieval between its current observation -- e.g. the sequence to the left of the grey box in~\cref{fig:dna-pr-system} -- and images from those past experiences.

This is done by nearest-neighbour lookups in the high-dimensional space of last-layer features generated by feature extractor networks (see~\cref{sec:backbones} for more detail).
For this we use a ``difference matrix'' on the right of~\cref{fig:dna-pr-system} which shows the embedding distances (Euclidean) between live and map features.
Localisation is successful if the nearest neighbour for a query embedding is a reference embedding which truly is close to the query location in physical space.
See~\cref{sec:metrics} for detail on performance assessment.

\subsection{Comparing visual observations \& experiences}

\begin{figure}[!h]
\centering
\includegraphics[width=0.4\textwidth]{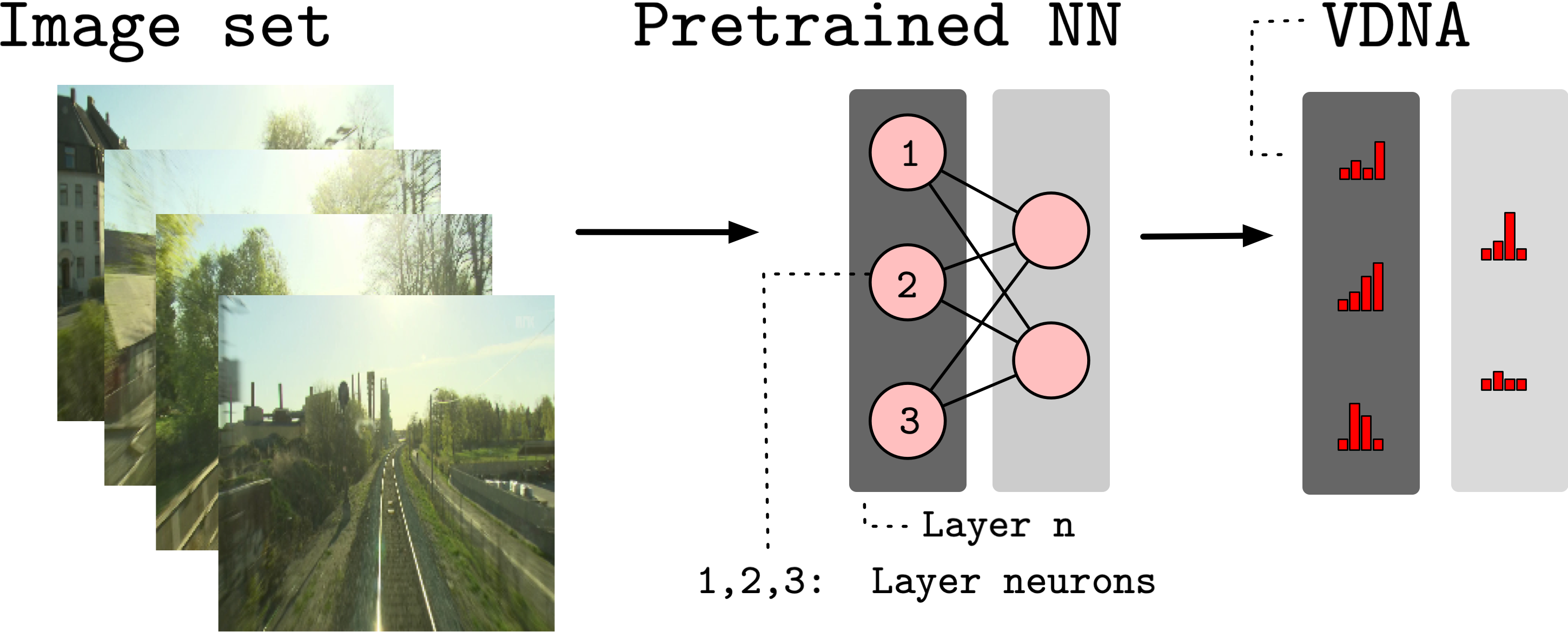}
\caption{
\textit{Visual DNA} is a collection of neuron-wise distributions of activation levels when either an image or an image set passes through the network in the forward direction.
From two \texttt{vdnas} -- e.g. for an image set in \texttt{Winter} and and an image set in \texttt{Summer}, we compute a similarity score by comparing histograms at each neuron by some distance function (see~\cref{sec:ranking_selecting_experiences}), and then averaging those scores.
}
\label{fig:vdna-overview}
\end{figure}

\cref{fig:vdna-overview} gives a basic overview of \textit{Visual DNA}, with more detail in~\cite{ramtoula2023cvpr}.
\texttt{Vdnas} are generated by passing images through a pre-trained and frozen feature extractor network, either convolutional, e.g. ResNet~\cite{he2016deep}, or based on transformers, e.g. ViT~\cite{dosovitskiy2020image}.
Distributions (specifically, histograms) are fit to neuron activations throughout the network.
The collection of such distributions is termed a ``Visual DNA'' (\texttt{vdna})
and is constant-sized regardless of the number of images it represents.
Thus, \texttt{vdnas} of \textit{entire experiences} (in the map) may be precomputed and cheaply compared to \texttt{vdnas} of live images.

\subsection{Warmup, selection, \& deployment}
\label{sec:warmup_etc}

For the \texttt{candidate experiences} in the map, \texttt{vdnas} are generated offline using all gathered images.
For example, in~\cref{fig:dna-pr-system} we have a \texttt{vdna} for the orange framed sequence and a \texttt{vdna} for the blue framed sequence, etc (both in the grey area).
We can use all map data since it has already been collected, and we do so to capture the experience's seasonal or illumination condition. 

Online, we would like to use this to inform real-time localisation.
As motived in~\cref{fig:spring-diff-mats}, single-experience selection either achieves the best performance or gets us closest to the performance of composite map experiences.
Thus, \texttt{vdnas} are generated with a smaller set of \texttt{warmup} images (e.g. the last few seconds of data under motion), shown by the red framed images in~\cref{fig:dna-pr-system}.
Details of the size of this \texttt{warmup} set differ for our two datasets (see \cref{sec:cross_season_dataset,sec:illumination_dataset}).

The closest \texttt{candidate experience} is then selected to localise the \texttt{live} data (see~\cref{sec:topological_localisation_system}).
In~\cref{fig:dna-pr-system} this is because the orange \texttt{vdna} was closer to the blue \texttt{vdna} (distances measured as per~\cref{sec:ranking_selecting_experiences}).

\subsection{Ranking \& selecting experiences}
\label{sec:ranking_selecting_experiences}

Equipped with \texttt{vdnas} of experiences and current observations, we can rank the experiences by \texttt{vdna} similarities, using a histogram distance function, the Earth-Mover's Distance (\texttt{emd})~\cite{rubner2000earth}.
This provides one distance per neuron, the full set of which (from the layer of interest) we average to measure the domain gap between observed and experience images.
Specifically, we compare the \texttt{vdna} of \texttt{warmup} to each \texttt{candidate experience} in~\cref{fig:dna-pr-system}, using \texttt{emd} to reduce each \texttt{vdna} comparison to a single scalar, which is smaller if the datasets are more similar -- if the domain gaps are less significant.
We hypothesise that \texttt{vdna} differences across image sets of the same route but different conditions will be correlated with place recognition performance.

\subsection{Backbones used}
\label{sec:backbones}

We investigate performance over two pretrained neural networks on both ranking and localising, including:
\begin{inparaenum}[(1)]
\item \texttt{CosPlace (Resnet101)}~\cite{berton2022rethinking}, as a model trained specifically for place recognition, the task at hand, which generalises well to different datasets, and
\item \texttt{Mugs (ViT-B/16)}~\cite{zhou2022mugs}, a self-supervised method focused on learning a general, multi-granular representation.
We use this model as recent work has demonstrated that general feature representations from pretrained self-supervised models are an excellent solution for universal visual place recognition~\cite{keetha2023anyloc}.
\end{inparaenum}
Neurons from the last layer of each model are used for both \texttt{vdna} comparisons and localisation.

\section{Experimental Setup}
\label{sec:experimental_setup}

To evaluate our approach, we measure the localisation performance of a query sequence when choosing only one experience to localise to, and find the cases in which some other map experience will have yielded better localisation performance, computing a ranking error which expresses this as a single number as the consequential drop in localisation performance. 

\subsection{Nordland dataset}
\label{sec:cross_season_dataset}

\begin{figure}[t]
\centering
\includegraphics[width=0.17\textwidth]{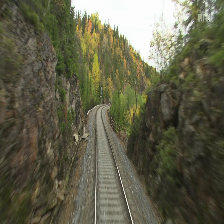}
\includegraphics[width=0.17\textwidth]{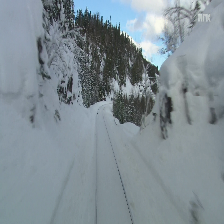}
\includegraphics[width=0.17\textwidth]{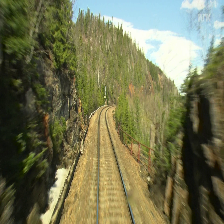}
\includegraphics[width=0.17\textwidth]{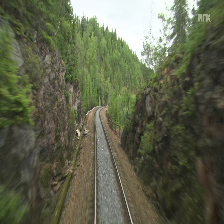}
\caption{\textit{Nordland} dataset samples: \texttt{Fall}, \texttt{Winter}, \texttt{Spring}, \texttt{Summer} (left to right).}
\label{fig:nordland_examples}
\vspace{-15pt}
\end{figure}

To investigate seasonal-semantic changes, we use the \textit{Nordland dataset}~\cite{sunderhauf2013we}, some samples of which are shown in~\cref{fig:nordland_examples}.
This consists of four train journeys in \texttt{Summer}, \texttt{Fall}, \texttt{Winter}, and \texttt{Spring} across Nordland in Norway.
As the train is on fixed tracks, this dataset does not feature any viewpoint variation.
Therefore, this dataset is ideal for isolating seasonal-semantic shifts.

We use the held-out, non-overlapping partitions from~\cite{olid2018single}, i.e. \texttt{test:1}, \texttt{test:2}, and \texttt{test:3}, each \texttt{test} sequence consisting of $1150$ images\footnote{See~\rurl{webdiis.unizar.es/~jmfacil/pr-nordland}} .
For example, we localise query frames from \texttt{Summer-test:1} to maps built from \texttt{Winter-test:1} or \texttt{Fall-test:1}, etc, while \texttt{test:2} queries are localised to \texttt{test:2} maps, etc. 

The \textit{Nordland} videos are exactly synchronised, so ground truth distances are taken as the difference in video frame number.
A positive localisation match is within $5$ frames, as used in~\cite{sunderhauf2015performance}.
The first $100$ frames of each \texttt{test} split are used as the \texttt{warmup} frames (\cref{sec:warmup_etc}).

\subsection{University Parks dataset}
\label{sec:illumination_dataset}

\begin{figure}
\centering
\vspace{-17.5pt}
\includegraphics[width=\textwidth]{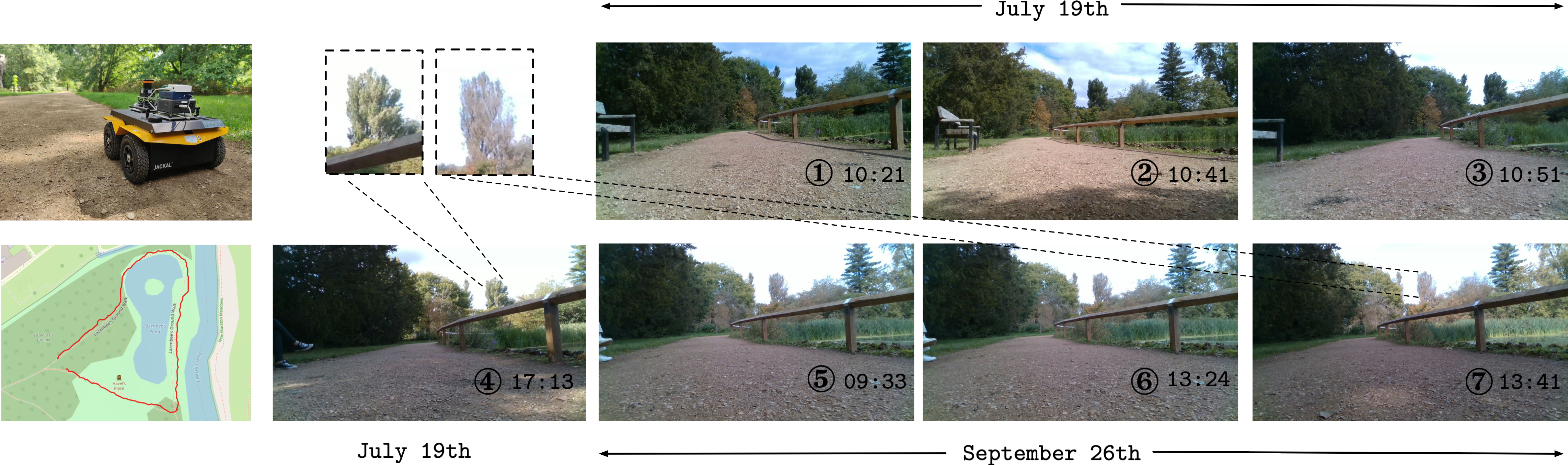}
\caption{
\textit{Top left: } \textit{Clearpath Jackal} platform with \textit{Intel RealSense} camera.
\textit{Bottom left: } GPS trace of route driven at the \textit{University Parks}.
\textit{Top row (right): } Sample images from three \texttt{July 19th} sequences collected during Summer, showing only illumination variation.
\textit{Bottom row (right): } Sample images from three \texttt{September 26th} sequences collected during Fall, showing consistent illumination on that day but seasonal variation versus \texttt{July 19th}.
As in~\cref{sec:cross_season_dataset}, one of (\ding{172},\ding{173},\ding{174}) is the query, and the others are candidate map experiences -- and this is cross-validated.
Similarly for the set (\ding{175},\ding{176},\ding{177},\ding{178}).
}
\label{fig:experiments}
\vspace{-30pt}
\end{figure}

We also perform practical experiments on a robot in an outdoor environment subject to lighting and seasonal variation, as shown in~\cref{fig:experiments}. 

We use a \textit{Clearpath Jackal UGV} equipped with an Intel \textit{RealSense D435} from which we capture RGB images for localisation and an integrated GPS to provide localisation ground truth and validate our predictions.
The platform is equipped with an \textit{Intel NUC8i7BEH} for onboard processing.

We deploy this robot in Oxford's \textit{University Parks}, a wooded area with varying tree cover.
We follow a trajectory several times, logging images and GPS.

Here, a match is considered good if it is within \SI{5}{\metre} of the query location.
We rely on the first \SI{45}{\second} of images as the \texttt{warmup} images.

\subsection{Measuring localisation performance}
\label{sec:metrics}

\Cref{fig:spring-diff-mats} shows examples of image-embedding difference matrices (see~\cref{sec:topological_localisation_system}).
The matrices have rows equal to the number of query images and columns equal to the number of reference images.
White regions mean that embeddings are distant from each other, while black regions mean that they are close.
There is a corresponding ground truth distance matrix.

Localisation performance is measured by \texttt{Recall@1}, the percentage of live queries for which the \textit{nearest} reference embedding is actually nearby in ground truth.
This corresponds to finding the index of the smallest element for each row of the distance matrix and confirming that the element at that location in the ground truth matrix is actually close to the query. 

\subsection{Optimal performance, multiple-experience localisation}
\label{sec:optimal}

\begin{figure}
\centering
\vspace{-25pt}
\includegraphics[width=\textwidth]{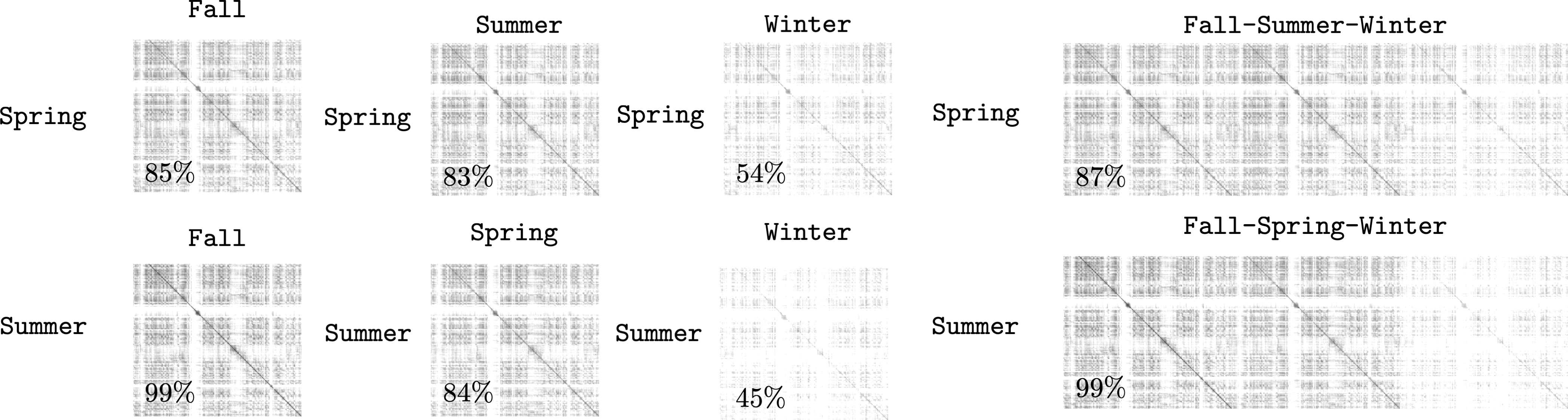}
\caption{\textit{Motivation for experience-selection}.
\textit{Top} shows the embedding distance difference matrices used for localisation between \texttt{Spring} as a query, and individually using \texttt{Fall}, \texttt{Summer}, and \texttt{Winter} as single-experience maps to localise to, as well as a map consisting of all three \texttt{Fall}, \texttt{Summer}, and \texttt{Winter}.
Here, a mix of experiences is important for best performance.
\textit{Top} does the same but, for \texttt{Summer} as query.
With more detail in~\cref{tab:1_gt}, the single best experience more often dominates performance even in a multi-experience setup, and so judicious experience selection is important.
}
\label{fig:spring-diff-mats}
\vspace{-15pt}
\end{figure}

Column-headers in \cref{tab:1_gt,tab:app_gt,tab:app_gt} indicate optimal performance when all reference experiences are used in a multiple-experience map.
The column data, instead, indicate performances when using as maps only single selected experience.
This is illustrated and motivated in~\cref{fig:spring-diff-mats}, where in one case several reference experiences would best explain the query data, but often there is one privileged experience which is best to use.

\subsection{Baselines}
\label{sec:baselines}

Related to \textit{Visual DNA}, the Fr\'echet Inception Distance (FID)~\cite{heusel2017gans} fits a multivariate Gaussian to the embeddings of all images from an \textit{InceptionV3}~\cite{berton2022rethinking} layer's feature space.
This is also suitable as an observation-to-experience comparison, and we compare it to \texttt{vdna} in~\cref{sec:results}.
However, rather than \textit{InceptionV3}, we perform Fr\'echet Distance (FD) on the backbones mentioned above (for fair comparison and to avoid biasing to \textit{ImageNet}~\cite{imagenet} classes~\cite{kynkaanniemi2022imagenetfid}).

As a further simple baseline, we also use the average pixel intensity of an image or average pixel intensity over images to rank experiences (importantly, never to perform localisation, which still relies on deep features).
Lastly, we also use two simple baselines.
Firstly, \texttt{candidate} experiences are selected randomly to match to.
Secondly, the composite of all experiences can be searched for localisation matches.
These simple baselines ensure that the task is not trivial and that discarding some experiences does not cause large performance drops.

\subsection{Measuring ranking errors}
\label{sec:exp:ranking}

\cref{tab:1_gt} shows an example of the ground truth ranking of experiences by actual \texttt{Recall@1} localisation performance.
\cref{tab:1_vdna,tab:1_fd,tab:1_pixel} then show the ranking of experiences by our proposed method (\cref{sec:ranking_selecting_experiences}) as well as baselines (\cref{sec:baselines}).
Ranking errors in~\cref{tab:1_vdna,tab:1_fd,tab:1_pixel} are orange/red by severity ($1$ or more incorrect positions as determined by~\cref{tab:1_gt}).

To summarise experience selection capability as a single number, ranking errors are weighted by~\cref{tab:1_gt} \texttt{Recall@1} discrepancy.
For example, FD swaps \texttt{Fall} and \texttt{Summer} (orange), and this is penalised as $|48.47\%-40.31\%|=8.16\%$) and contributes to the average\footnote{Over $72=12\times3\times2$ experiments ($12$ experience-pairs, $3$ spatial splits, $2$ backbones)} in~\cref{tab:avg_errors}.

\section{Results}
\label{sec:results}

{\bf Nordland dataset:}
Consider~\cref{tab:nordland_results_full} in which~\cref{tab:avg_errors} averages ranking errors (\cref{sec:exp:ranking}) over all \textit{Nordland} partitions (\cref{sec:cross_season_dataset}).
Here, the benefit of dataset-to-dataset comparison tools in this application becomes evident, far outperforming the pixel intensity baseline.
Consider that the \textit{Nordland} dataset exhibits seasonal variation, resulting in changed semantic content (bare trees, snow fall, lusher vegetation, etc). 
Thus, DNA and FD better capture the domain shift by measuring statistical variation in the semantically responsive layers.
Pixel intensity is not well-suited to this sort of change and \textit{Visual DNA} is the best predictor of deployment-time performance -- better measuring extreme seasonal domain shift.

For example, \cref{tab:1_gt,tab:1_fd,tab:1_pixel,tab:1_vdna} show the ranking for the \texttt{test:2} partition, as an example.
In~\cref{tab:1_gt}, we see that if \texttt{Winter} were the live experience, then the best experience to localise to is \texttt{Spring}, followed by \texttt{Summer}, and then \texttt{Fall}.
FD makes a $8.16\%$ mistake in~\cref{tab:1_fd}, as we have given as an example in~\cref{sec:exp:ranking}.
\textit{Visual DNA}, in~\cref{tab:1_vdna}, does not make that same mistake.
We also observe in~\cref{tab:1_pixel} the predicted experience rank by the pixel intensity baseline, which makes the same mistake as FD for the \texttt{Winter} query, but also makes more serious mistakes with \texttt{Spring} as query -- i.e. choosing \texttt{Winter} as the best experience to localise to, whereas in fact it is the worst (red).
Additionally, selecting experiences by \textit{Visual DNA} always selects the best experience, which has similar (e.g. $98.98\%$ for \texttt{Fall-vs-Summer}) or close performance to the best combination from using all reference experiences (e.g. $87.76\%$ vs $85.20\%$ for \texttt{Spring}) and has the significant benefit of meaning that only one experience is localised against at all -- saving compute effort.

\begin{table}
\centering
\begin{subtable}[h]{0.48\textwidth}
\resizebox{\columnwidth}{!}{
\begin{tabular}{l  l  l  l  l}\toprule
\multicolumn{1}{c}{} & \multicolumn{4}{c}{\textit{Query / \texttt{Recall@1} with access to all references }} \\
\cmidrule(lr){2-5}
& Fall / 98.98 & Winter / 61.22 & Summer / 99.49 & Spring / 87.76 \\
\cmidrule(lr){2-2}\cmidrule(lr){3-3}\cmidrule(lr){4-4}\cmidrule(lr){5-5}
\multirow{4}{*}{\rotatebox{90}{\parbox{.5cm}{\centering \textit{Reference}}}} 
& 98.98 - Summer & 61.22 - Spring & 99.49 - Fall & 85.20 - Fall\\
& 84.69 - Spring & 48.47 - Summer & 84.69 - Spring & 83.16 - Summer\\
& 41.84 - Winter & 40.31 - Fall & 45.92 - Winter & 54.08 - Winter\\
\bottomrule
\end{tabular}
}
\caption{Localisation rank by \texttt{Recall@1} (\%), $\downarrow$.\label{tab:1_gt}}
\end{subtable}
%
\begin{subtable}[h]{0.48\textwidth}
\resizebox{\columnwidth}{!}{
\begin{tabular}{l  l  l  l  l}\toprule
\multicolumn{1}{c}{} & \multicolumn{4}{c}{\textit{Query}} \\
\cmidrule(lr){2-5}
& Fall & Winter & Summer & Spring \\
\cmidrule(lr){2-2}\cmidrule(lr){3-3}\cmidrule(lr){4-4}\cmidrule(lr){5-5}
\multirow{4}{*}{\rotatebox{90}{\parbox{.5cm}{\centering \textit{Reference}}}} 
& 5.68 - Summer & 9.11 - Spring & 5.41 - Fall & 8.05 - Fall\\
& 6.71 - Spring & 11.22 - Summer & 6.67 - Spring & 8.23 - Summer\\
& 11.53 - Winter & 11.27 - Fall & 10.94 - Winter & 9.63 - Winter\\
\bottomrule
\end{tabular}
}
\caption{Experience rank by DNA (Ours), $\uparrow$.\label{tab:1_vdna}}
\end{subtable}
\begin{subtable}[h]{0.48\textwidth}
\resizebox{\columnwidth}{!}{
\begin{tabular}{l  l  l  l  l}\toprule
\multicolumn{1}{c}{} & \multicolumn{4}{c}{\textit{Query}} \\
\cmidrule(lr){2-5}
& Fall & Winter & Summer & Spring \\
\cmidrule(lr){2-2}\cmidrule(lr){3-3}\cmidrule(lr){4-4}\cmidrule(lr){5-5}
\multirow{4}{*}{\rotatebox{90}{\parbox{.5cm}{\centering \textit{Reference}}}} 
& 0.264 - Summer & 0.435 - Spring & 0.252 - Fall & 0.371 - Fall\\
& 0.308 - Spring & \cellcolor{orange!25}0.566 - Fall & 0.306 - Spring & 0.384 - Summer\\
& 0.615 - Winter & \cellcolor{orange!25}0.568 - Summer & 0.583 - Winter & 0.476 - Winter\\
\bottomrule
\end{tabular}
}
\caption{Experience rank by FD, $\uparrow$.\label{tab:1_fd}}
\end{subtable}
\begin{subtable}[h]{0.48\textwidth}
\resizebox{\columnwidth}{!}{
\begin{tabular}{l  l  l  l  l}\toprule
\multicolumn{1}{c}{} & \multicolumn{4}{c}{\textit{Query}} \\
\cmidrule(lr){2-5}
& Fall & Winter & Summer & Spring \\
\cmidrule(lr){2-2}\cmidrule(lr){3-3}\cmidrule(lr){4-4}\cmidrule(lr){5-5}
\multirow{4}{*}{\rotatebox{90}{\parbox{.5cm}{\centering \textit{Reference}}}} 
& 9.35 - Summer & 5.38 - Spring & 7.91 - Fall & \cellcolor{red!25}1.15 - Winter\\
& 15.45 - Spring & \cellcolor{orange!25}27.50 - Fall & 14.21 - Spring & \cellcolor{orange!25}26.48 - Fall\\
& 20.97 - Winter & \cellcolor{orange!25}30.19 - Summer & 19.73 - Winter & \cellcolor{orange!25}29.17 - Summer\\
\bottomrule
\end{tabular}
}
\caption{Experience rank by pixel intensity, $\uparrow$.\label{tab:1_pixel}}
\end{subtable}

\begin{subtable}{0.8\textwidth}
\centering
\resizebox{0.675\columnwidth}{!}{
\begin{tabular}{c|c|c|c|c|c}
Backbone & Random & Pixel Intensity & FD & DNA (Ours) \\
\hline
\texttt{CosPlace (Resnet101)} & 9.83\% & 4.71\% & 0.53\% & \cellcolor{gray!25}\textbf{0.17\%} \\
\texttt{Mugs (ViT-B/16)} & 13.69\% & 7.92\% & 0.35\% & \cellcolor{gray!25}\textbf{0.29\%} \\
\hline
\textbf{Average} & 11.76\% & 6.31\% & 0.44\% & \cellcolor{gray!25}\textbf{0.23\%}
\end{tabular}
}
\caption{Ranking errors averaged over networks/splits.\label{tab:avg_errors}}
\end{subtable}
\caption{
Detailed experience ordering for a given query on the \textit{Nordland dataset}. 
\Cref{tab:1_gt,tab:1_vdna,tab:1_fd,tab:1_pixel} show example results for \texttt{cosplace\_resnet101\_128} and the \texttt{test:2} split, while~\cref{tab:avg_errors} averages errors over \textit{all} networks and dataset splits.
\texttt{Random} is the average of all permutations of the \texttt{candidate experiences} list.
}
\label{tab:nordland_results_full}
\vspace{-30pt}
\end{table}

Finally, in~\cref{tab:3}, we investigate a situation where our system does not do perfectly -- this time over the \texttt{test:1} section.
\textit{Visual DNA} makes the same sort of mistake as just discussed in \texttt{Winter}, swapping the rank of \texttt{Fall} and \texttt{Summer}.
However, in this case the actual localisation performances for \texttt{Fall} and \texttt{Summer} are very close to each other, both approximately \SI{89}{\percent} (indistinguishable to three decimal places) whereas in \texttt{test:2} (\cref{tab:1_gt}) they differed by \SI{8}{\percent}.
Therefore, if our experience selection mechanism confuses these two experiences, it would not have serious consequences for localisation outcome.
\textit{Visual DNA} also makes a mistake in \texttt{Spring}, swapping \texttt{Summer} and \texttt{Fall} -- but, with only an approximately \SIrange{1}{2}{\percent} consequence in localisation performance.
FD makes a more serious error, swapping \texttt{Winter} and \texttt{Fall} with an approximately \SI{11}{\percent} consequence in localisation performance -- with worse results.

\begin{table}[!h]
\centering
\vspace{-15pt}
\begin{subtable}[h]{0.48\textwidth}
\resizebox{\columnwidth}{!}{
\begin{tabular}{l  l  l  l  l}\toprule
\multicolumn{1}{c}{} & \multicolumn{4}{c}{\textit{Query / \texttt{Recall@1} with access to all references }} \\
\cmidrule(lr){2-5}
& Fall / 100.00 & Winter / 93.81 & Summer / 100.00 & Spring / 94.76 \\
\cmidrule(lr){2-2}\cmidrule(lr){3-3}\cmidrule(lr){4-4}\cmidrule(lr){5-5}
\multirow{4}{*}{\rotatebox{90}{\parbox{.5cm}{\centering \textit{Reference}}}} 
& 100.000 - Summer & 90.952 - Spring & 100.000 - Fall & 94.286 - Fall\\
& 91.429 - Spring & 89.048 - Summer & 92.381 - Spring & 92.857 - Summer\\
& 77.143 - Winter & 89.048 - Fall & 79.048 - Winter & 83.333 - Winter\\
\bottomrule
\end{tabular}
}
\caption{Localisation rank by \texttt{Recall@1} (\%), $\downarrow$.}
\label{tab:app_gt}
\end{subtable}
%
\begin{subtable}[h]{0.48\textwidth}
\resizebox{\columnwidth}{!}{
\begin{tabular}{l  l  l  l  l}\toprule
\multicolumn{1}{c}{} & \multicolumn{4}{c}{\textit{Query}} \\
\cmidrule(lr){2-5}
& Fall & Winter & Summer & Spring \\
\cmidrule(lr){2-2}\cmidrule(lr){3-3}\cmidrule(lr){4-4}\cmidrule(lr){5-5}
\multirow{4}{*}{\rotatebox{90}{\parbox{.5cm}{\centering \textit{Reference}}}} 
& 9.134 - Summer & 10.867 - Spring & 9.041 - Fall & \cellcolor{orange!25}9.951 - Summer\\
& 9.724 - Spring & \cellcolor{orange!25}11.311 - Fall & 9.480 - Spring & \cellcolor{orange!25}9.969 - Fall\\
& 9.781 - Winter & \cellcolor{orange!25}11.450 - Summer & 9.939 - Winter & 10.036 - Winter\\
\bottomrule
\end{tabular}
}
\caption{Experience rank by DNA (Ours), $\uparrow$.}
\end{subtable}
\begin{subtable}[h]{0.48\textwidth}
\resizebox{\columnwidth}{!}{
\begin{tabular}{l  l  l  l  l}\toprule
\multicolumn{1}{c}{} & \multicolumn{4}{c}{\textit{Query}} \\
\cmidrule(lr){2-5}
& Fall & Winter & Summer & Spring \\
\cmidrule(lr){2-2}\cmidrule(lr){3-3}\cmidrule(lr){4-4}\cmidrule(lr){5-5}
\multirow{4}{*}{\rotatebox{90}{\parbox{.5cm}{\centering \textit{Reference}}}} 
& 0.671 - Summer & 0.767 - Spring & 0.656 - Fall & \cellcolor{red!25}0.722 - Winter\\
& 0.694 - Spring & \cellcolor{orange!25}0.809 - Fall & 0.676 - Spring & 0.734 - Summer\\
& 0.727 - Winter & \cellcolor{orange!25}0.824 - Summer & 0.724 - Winter & \cellcolor{red!25}0.742 - Fall\\
\bottomrule
\end{tabular}
}
\caption{Experience rank by FD, $\uparrow$.}
\end{subtable}
%
\begin{subtable}[h]{0.48\textwidth}
\resizebox{\columnwidth}{!}{
\begin{tabular}{l  l  l  l  l}\toprule
\multicolumn{1}{c}{} & \multicolumn{4}{c}{\textit{Query}} \\
\cmidrule(lr){2-5}
& Fall & Winter & Summer & Spring \\
\cmidrule(lr){2-2}\cmidrule(lr){3-3}\cmidrule(lr){4-4}\cmidrule(lr){5-5}
\multirow{4}{*}{\rotatebox{90}{\parbox{.5cm}{\centering \textit{Reference}}}} 
& 21.705 - Summer & \cellcolor{red!25}6.151 - Fall & 15.081 - Fall & \cellcolor{red!25}21.595 - Winter\\
& \cellcolor{orange!25}38.738 - Winter & 7.714 - Summer & \cellcolor{orange!25}15.817 - Winter & 38.628 - Summer\\
& \cellcolor{orange!25}66.083 - Spring & \cellcolor{red!25}52.092 - Spring & \cellcolor{orange!25}43.162 - Spring & \cellcolor{red!25}52.493 - Fall\\
\bottomrule
\end{tabular}
}
\caption{Experience rank by pixel intensity, $\uparrow$.}
\end{subtable}
\caption{
\texttt{CosPlace (Resnet101)} over \texttt{test:1} of the \textit{Nordland dataset}.
}
\label{tab:3}
\vspace{-30pt}
\end{table}

{\bf University Parks dataset:}
We perform two experiments over the data in our \textit{University Parks} deployment.
Firstly, \cref{tab:parks_a} and \cref{tab:parks_c} consider three experiences from \texttt{July 19th} in~\cref{fig:experiments}, i.e. with varying illumination and no seasonal change.
Secondly, \cref{tab:parks_b} and \cref{tab:parks_d} considers four experiences where three are from \texttt{September 26th} in~\cref{fig:experiments} and do not exhibit illumination variance amongst themselves but are from a different season than one experience from \texttt{July 19th} which is also included.
The localisation ranks in \cref{tab:parks_a} and \cref{tab:parks_b} are given as examples for \texttt{CosPlace (Resnet101)} while the ranking errors are aggregated over \texttt{CosPlace (Resnet101)} and \texttt{Mugs (ViT-B/16)}.

Interestingly, for the \texttt{July 19th} experiments, the Pixel Intensity baseline performs best (with no ranking errors in \cref{tab:parks_a} for \texttt{CosPlace (Resnet101)}), while FD and DNA perform equivalently -- \textit{at best} matching pixel intensity for \texttt{Mugs (ViT-B/16)} in \cref{tab:parks_c}.
Still, all three approaches perform very well, producing average experience rank errors of $1.78\%$ at most.
It is important to note that under these experimental conditions illumination does not correspond with time of day, with e.g. \texttt{10:21} and \texttt{10:51} more overcast in contrast to more direct sun during \texttt{10:41}, as is clear in~\cref{fig:experiments}.
The minor dominance of pixel intensity under these conditions is sensible, as we are only seeing time-of-day and thus illumination changes which will affect object edges, textures, etc, low-level image changes for which neural networks are more responsive at earlier layers~\cite{ou2019moving}, which are not used here.
Pixel Intensity directly measures the observed changes in the environment, which are sufficient with the limited variations in this setting, but struggles more with higher-level changes.

Indeed, for \texttt{September 26th}, Pixel Intensity performs poorly and FD/DNA neural dataset-to-dataset comparisons again outperform it.
This is again in the face of seasonal variation (e.g. see the browning tree in~\cref{fig:experiments}), for which Pixel Intensity cannot provide a measure of high-level variation.
Interestingly, for this experiment, FD performs best, which could be linked to how FD and DNA perform with different number of images to represent compared datasets.
Also note that \texttt{vdna} matches performance with FD for \texttt{CosPlace (Resnet101)}, a network trained specifically for place recognition, in~\cref{tab:parks_d}. However, the best results for both comparison techniques are obtained using \texttt{Mugs (ViT-B/16)}.
This is the opposite of what we observed on the \textit{Nordland dataset}, confirming that self-supervised networks can compete with models trained specifically on place recognition when considering different domains.

\begin{table}
\centering
\begin{subtable}[h]{0.45\textwidth}
\resizebox{\columnwidth}{!}{
\begin{tabular}{l  @{\hskip 0.15in} l  @{\hskip 0.15in} l  @{\hskip 0.15in} l}\toprule
\multicolumn{1}{c}{} & \multicolumn{3}{c}{\textit{Query}} \\
\cmidrule(lr){2-4}
& 10:21 19/07 & 10:41 19/07 & 10:51 19/07 \\
\cmidrule(lr){2-2}\cmidrule(lr){3-3}\cmidrule(lr){4-4}
\multirow{4}{*}{\rotatebox{90}{\parbox{.1cm}{\centering \textit{Reference}}}} 
& 0.436 - 10:51  & 0.630 - 10:51 & 0.634 - 10:41\\
& 0.418 - 10:41 & 0.449 - 10:21 & 0.425 - 10:21
\\\bottomrule
\end{tabular}
}
\caption{\texttt{Recall@1} rank $\downarrow$.\label{tab:parks_a}}
\end{subtable}
\begin{subtable}[h]{0.5\textwidth}
\resizebox{\textwidth}{!}{
\begin{tabular}{l  @{\hskip 0.15in}l@{\hskip 0.15in}  l @{\hskip 0.15in} l @{\hskip 0.15in} l}\toprule
\multicolumn{1}{c}{} & \multicolumn{4}{c}{\textit{Query}} \\
\cmidrule(lr){2-5}
& 17:13 19/07 & 09:33 26/09 & 13:24 26/09 & 13:41 26/09 \\
\cmidrule(lr){2-2}\cmidrule(lr){3-3}\cmidrule(lr){4-4}\cmidrule(lr){5-5}
\multirow{4}{*}{\rotatebox{90}{\parbox{.1cm}{\centering \textit{Reference}}}} 
& 37.41 - 09:33 & 56.36 - 13:24 & 72.06 - 13:41 & 70.86 - 13:24 \\
& 35.79 - 13:41 & 55.84 - 13:41 & 64.05 - 09:33 & 63.37 - 09:33 \\
& 30.30 - 13:24 & 19.59 - 17:13 & 19.78 - 17:13 & 29.99 - 17:13 \\
\bottomrule
\end{tabular}
}
\caption{\texttt{Recall@1} rank $\downarrow$.\label{tab:parks_b}}
\end{subtable}
\begin{subtable}[h]{0.49\textwidth}
\resizebox{\columnwidth}{!}{
\begin{tabular}{c|c|c|c}
Backbone & Pixel Intensity & FD & DNA (Ours) \\
\hline
\texttt{CosPlace (Resnet101)} & \cellcolor{gray!25}\textbf{0.00\%} & 0.30\% & 0.30\% \\
\texttt{Mugs (ViT-B/16)} & \cellcolor{gray!25}\textbf{1.78\%} & \cellcolor{gray!25}\textbf{1.78\%} & \cellcolor{gray!25}\textbf{1.78\%} \\
\hline
\textbf{Average} & \cellcolor{gray!25}\textbf{0.89\%} & 1.04\% & 1.04\% \\
\end{tabular}
}
\caption{Average experience rank errors.\label{tab:parks_c}}
\end{subtable}
\begin{subtable}[h]{0.49\textwidth}
\resizebox{\textwidth}{!}{
\begin{tabular}{c|c|c|c}
Backbone & Pixel Intensity & FD & DNA (Ours) \\
\hline
\texttt{CosPlace (Resnet101)} & 14.20\% & \cellcolor{gray!25}\textbf{6.25}\% & \cellcolor{gray!25}\textbf{6.25}\% \\
\texttt{Mugs (ViT-B/16)} & 12.92\% & \cellcolor{gray!25}\textbf{1.18\%} & 3.97\% \\
\hline
\textbf{Average} & 13.56\% & \cellcolor{gray!25}\textbf{3.72\%} & 5.11\% \\
\end{tabular}
}
\caption{Average experience rank errors.\label{tab:parks_d}}
\end{subtable}
\caption{
\textit{University Parks experiments} for \subref{tab:parks_a}, \subref{tab:parks_c} \texttt{July 19th} and \subref{tab:parks_b}, \subref{tab:parks_d} \texttt{September 26th} experience groups from \cref{fig:experiments}.
Note that in \subref{tab:parks_b} to save space we list only the time of day, but in actual fact \texttt{17:13} is from \texttt{July 19th} while the other columns are from \texttt{September 26th}. 
}
\vspace{-25pt}
\end{table}

\section{Conclusion}
\label{sec:conclusion}

Overall, we have presented a new approach to characterising dataset-level differences due to appearance change.
For cases of extreme seasonal change, our proposed measure is more highly correlated with actual localisation performance than several baselines.
Thus, \textit{Visual DNA} is a good candidate for dataset-to-dataset similarity measurement to predict experience rank in visual localisation.

These experimental results thus open new lines of investigation, including (1) pruning network neurons sensitive to appearance variation for improved and customised dataset-to-dataset comparison techniques, (2) using deep neural dataset-to-dataset similarity directly to perform sequential place recognition.

\section*{Acknowledgements}

The authors are grateful to David Williams for his assistance in collecting data.

\bibliographystyle{IEEEtran}
\bibliography{biblio}

\end{document}